# General Vector Machine

Hong Zhao (zhaoh@xmu.edu.cn)

Department of Physics, Xiamen University

The support vector machine (SVM) is an important class of learning machines for function approach, pattern recognition, and time-serious prediction, etc. It maps samples into the feature space by so-called support vectors of selected samples, and then feature vectors are separated by maximum margin hyperplane. The present paper presents the general vector machine (GVM) to replace the SVM. The support vectors are replaced by general project vectors selected from the usual vector space, and a Monte Carlo (MC) algorithm is developed to find the general vectors. The general project vectors improves the feature-extraction ability, and the MC algorithm can control the width of the separation margin of the hyperplane. By controlling the separation margin, we show that the maximum margin hyperplane can usually induce the overlearning, and the best learning machine is achieved with a proper separation margin. Applications in function approach, pattern recognition, and classification indicate that the developed method is very successful, particularly for small-set training problems. Additionally, our algorithm may induce some particular applications, such as for the transductive inference.

## 1. Introduction

In essence, a learning machine is a high-dimensional map from input $x \in R^M$ to output $y \in R^L$, $y = \Phi(\Lambda, x)$, where $\Lambda$ represents the parameter set. The goal of designing such a map is to find a parameter set $\Lambda$ in the condition of a given sample set $(x^\mu, y^\mu, \mu = 1, ..., P)$ of $P$ samples, which guarantees the map not only correctly response to the sample set, but also correctly response to the real set represented by the samples. Generalizing the knowledge learned from the limited samples to the real set is the key performance of a leaning machine. This ability is determined by the prior knowledge being incorporated into the learning machine. The prior knowledge is the information about the learning task which is available in addition to the training samples. A method for designing a learning machine is therefore usually composed of two parts, i.e., an algorithm for training the learning machine response to samples and strategies for gaining the generalization ability.

There are two fundamental methods for designing multilayer learning machines, i.e., the back-propagation (BP) method [1-3] and the support vector machine (SVM) method [4-5]. Both methods have achieved great success in various applications ranging from the traditional domain of function approach, pattern recognition and time series prediction to an increasingly wide variety of biological applications [1-7]. The BP method employs a set of deterministic equations to calculate the parameter set in an iterative manner to train the machine response to samples. To gain the generalization ability, it applies the 'Occam razor principle': the entities should not be multiplied beyond the necessity. Based on this principle, the best machine should have as smaller as possible size. However, it is already clear that smaller machine may not always give the best performance [4]. In addition, the BP algorithm is experience dependent, particularly for choosing a proper learning rate.

The SVM method is a great progress in designing learning machines. It gives up the Occam razor, and follows the structural risk minimization

(SRM) principle of statistical learning theory [4] to gain the generalization ability. Based on the SRM principle, the machine with the smallest Vapnik-Chervonenkis (VC) dimension instead of the smallest size is supposed to be the best. In more detail, the SVM method maps the input vectors of samples into the high-dimensional feature space by so-called support vectors, and then the feature vectors are separated with the maximum margin hyperplane calculated following the linear optimization algorithm. However, fundamentally, whether the structural risk can generally led to the best machine, particularly, whether the maximum margin hyperplane can be applied as a quantitative criterion of the best machine, are unknown[4,8]. Technically, how to calculate the VC dimension and how to choose the kernel function are also open problems [8]. In addition, support vectors are input vectors of special samples chosen from the training samples, which may result in serious restriction when only a small training set is available. Another problem is that the SVM is originally developed for binary decision problems. Though there are some efforts to extend the method to multi-classification problems [9], it seems that the performance of such a machine is still poorer than the bi-classification SVMs [10]. The SVM method thus needs to be further developed.

In this paper, we present a new method to design learning machines with input, hidden, and output layers. To train the learning machine responses to samples, we develop a Monte Carlo (MC) algorithm. In fact, Hagan, Demuth and Beale have pointed out that randomly searches for suitable weights may be a possible way. It was however abandoned because they did not believe it is practicable [2] because of the computational difficulty. They thus turn to develop the BP algorithm. We recur this idea since firstly it is available for finding solutions of optimization problems with complex restrictions as demanded by our generalization strategies. Secondly and essentially, the speed of nowadays' computers have greatly improved. Another equally important factor is that our algorithm evolves only a small part of the learning machine. At each round of adaptation, we adapt only one parameter if it does not make the performance bad, instead of developing the entire system. Due to this reason, the training time is acceptable for usual applications. As a MC algorithm, it has high flexibility, and is applicable for various neuron transfer functions and cost functions, and for either continuous or discontinuous, even discrete system parameters. Using this algorithm, we can directly search for the proper project vectors in the usual vector space. We thus call the learning machine designed by our method the general vector machine (GVM).

For pursing the generalization ability, we classify the prior knowledge into the common and problem-dependent classes, and suggest corresponding strategies to maximally integrate them into the learning machine. Objects with small difference in features should belong to the same class may be the most common prior knowledge for human beings to generalize experiences. To incorporate this kind of prior knowledge, a learning machine should be insensitive to small input changes, i.e., should have small input-output sensitivity. Mathematically, the amplitudes of derivatives, usually the second-order derivatives, define the structural risk and measure the input-output sensitivity of functions. Note that the concept of structural risk employed in SVM is not identical to such a general mathematical concept. Minimizing such a structural risk is indeed applied as a basic principle in the learning problem. However, in the case of multilayer learning machines, as a complex function of the system parameters and input vectors, the structural risk is quite difficult to be calculated, let alone finding its solution of minimum. In the SVM method, the separating margin in the feature space between different classes is maximized to decrease the input-output sensitivity. This strategy avoids the direct calculation of the structural risk. The problem is that extremely decreasing the structural risk could not always lead to the best machine, as our application examples will indicate. Another kind of the common prior knowledge is: learning machines supervised under the same training method with the same training set should have as small as possible output uncertainty. This is a basic requirement for a recognition system. It should has a sufficient degree of stability on small parameter changes. Otherwise, the system is lack of reliability. For example, the brain is remarkably robust; it does not stop working just because a few cells die.

We apply the design risk minimization (DRM) strategy -- Learning machines with smaller design risk are better ones, as a basic principle to maximally incorporate the common prior knowledge. The design risk is defined to be the dispersion degree of the outputs of learning machines designed using the same training set. It is just the error bar usually used to indicate the precision of the experimental data. Minimizing it minimizes the uncertainty of the outputs of different learning machine designed by the same algorithm. Furthermore, on one hand, small design risk means small input-output sensitivity. The design risk can thus impose restrictions on the structural risk. On the other hand, the DRM principle is not equivalent to the input-output sensitivity. The DRM can approach the minimum of the real risk better than the input-output sensitivity minimization strategy. In the case of function approach and smoothing, we will show that the minimum of the design risk is usually consistent to that of the real risk, while that of the structural risk has a degree of divergence from the real-risk minimum. For pattern recognition and objective classification, this is also true when the real patterns can be considered as the random variations of the training samples. We will explain why the DRM can achieve a better result and why the SRM may induce the divergence. However, there is no further rigid proof to this problem. Indeed, this is also the case for the SRM principle as well as the Occam razor principle, for which there is no rigid proof to guarantee the convergence to the real risk minimum when a finite sample set is available.

A particular set of samples has its special background knowledge. For the pattern recognition, for example, the patterns may have special geometric symmetry. In this case, extremely maximizing either the structural or the design risks may result the deviation from the nature geometry, and thus decreases the correct rate of recognition. The knowledge about the geometric symmetry of patterns is a kind of typical problem-dependent prior knowledge. Besides, the physical interpretation of input vectors of samples may also involve the problem-dependent prior knowledge, such as the knowledge behind the physiology and biochemistry indexes of the medical inspection. The problem-dependent prior knowledge presents in function approach and smoothing, too, such as the knowledge about the goal function. To maximally incorporate the problem-dependent prior knowledge, one needs to apply individualized strategies, including proper pretreatment of sample input vectors, using proper neuron transfer functions and cost functions, etc..

To search for the minimum of the design risk effectively, we introduce a set of control parameters. This set includes parameters specifying the ranges of the weights connecting different neurons, coefficients of neural transfer functions, and biases of neurons. The number of hidden-layer neurons and the width of the separation margin are also included. The control of the structural risk of the learning machine is approached by control the structural risk of individual neurons.

Using the design-risk-control strategy, the goal is to find the best control parameter set instead of finding the best learning machine. We introduce two performance indexes, the design risk and the average correct rate on the test set (or on a spurious test set), to identify the best control parameter set. In more detail, for function approach and smoothing, the design risk is applied to uniquely define the best control parameter set since we will demonstrate that the design risk gives a good estimate to the real risk. In this case, the design program is stopped when the design risk minimum is approached. For other problems, we apply the average correct rate on the test set as the dominant performance index. We show that for pattern recognition and objective classification, minimizing the design risk usually gives the highest average correct rate when real patterns can be considered as random variations of sample patterns. In this case, the two indexes are consistent with each other. When the problem-dependent prior knowledge is involved, however, the two indexes may be inconsistent with each other. In this case, the best control parameter set should be decided as a balance between the two performance indexes. In more details, if the design risk has become acceptable we identify the control parameter set with the maximum average correct rate to be the best control parameter set. If it is still too big, we have to searching alone the direction that the design risk decreases continuously till the risk is acceptable, though in that case the average correct rate become smaller.

The MC algorithm provides a nature way to calculate the design risk as well as the average correct rate, since for a given training set one can design a number of GVMs at a fixed control parameter set. Each round of training starts by setting all the system parameters to random numbers in their available ranges. The initial system parameters are independent and identically distributed, and the training is proceeded by MC adoptions.

These GVMs are thus statistically identical, and every GVM at the best control parameter set should have the same anticipation performance for the real set and can be equally applied as the learning machine to performing the task. This is different from previous methods which selecting the learning machine with the best performance on the test set to be the performing system, in which case one cannot guarantee that it also has the best performance for real set.

We can further construct the performing system by combining a sufficient number of GVMs designed at the same control parameter set and with the same training samples. We call it the joint GVM (J-GVM). The J-GVM can dramatically decrease the design risk as well as the structural risk. Moreover, the J-GVM may achieves a good balance between the goal of maximally extracting the feature of input vectors and the goal of minimizing the risks. It can therefore remarkably improve the generalization ability for small training-set problems, as will be shown by our examples. The idea of J-GVM is similar to the ensemble method [11]. The difference is that, besides being designed under the supervision of the DRM strategy, the GVMs which are used to construct the J-GVM are trained by the same training sample set.

The rest of the paper is managed as follows. In the next section we introduce the architecture of the GVM. We shall emphasize the difference from that of the SVM. In section 3 we present the MC algorithm for training a GVM to response to samples. Section 4 is contributed to introduce the idea of controlling the structural risk of the learning machine by controlling that of single neurons. The main control parameters are introduced in this section. Section 5 presents the DRM principle. Why minimizing the design risk can approach the best fitting in function approach and smoothing is explained. Section 6 introduces several strategies for maximally incorporating the problem-dependent prior knowledge. Section 7 constructs the J-GVM. The next three sections are application examples, including the function approach and smoothing, pattern recognition and classification respectively. Function approach examples show the consistence between the design risk minimum and the best fitting. The fitting precision outperforms the SVM method as well as the usual spline algorithm. Pattern recognition is performed on the famous MNIST set of handwritten digits [12]. Our purpose is to show how to perform this kind of task using a GVM or a J-GVM. We focus mainly on the case of using small training set. The recognition rate achieved using all the training samples is also shown for the comparison purpose. By directly using the normalized gray-scale images without special preprocessing so as to fairly compare the algorithms, we obtain recognition correct rate beyond the corresponding records by using the BP neural network, the SVM, and even the complex learning system supervised by the deep-learning method. The classification is performed on the Wisconsin breast cancer database [13-15]. This example fully reveal the advantage of our method on application to small-training set. The last section is to summarize the main ideas and results. A particular application, 'washing' out the bad samples, of our method is demonstrated in the end of this section.

## 2. The model

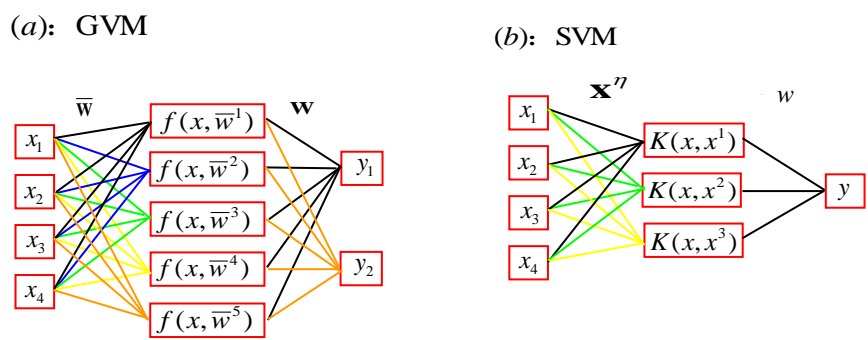

Figure 1. The architecture of a GVM (a) and a typical SVM (b).

We study the three-layer learning machine composed by input, hidden and output layers. The numbers of neurons in the input, hidden, and output layers are N, M, and L correspondingly. The dynamics is given by the following formula,

Hidden layer:

$$\bar{y}_i = f_i(\beta_i \bar{h}_i), \bar{h}_i = \sum_{j=1}^{M} \bar{w}_{ij} x_j - b_i, i = 1, ..., N \qquad (1)$$

Output layer:

$$y_l = h_l, h_l = \sum_{i=1}^{N} w_{li} \bar{y}_i, l = 1, ..., L \qquad (2)$$

where $\bar{y}_i, f_i, \bar{h}_i, \beta_i, b_i$ respectively represent the output, neuron transfer function, local field, transfer function coefficient, and bias of the $i$th neuron in the hidden layer. Here $x_j$ is the $j$th component of an input vector, $\bar{w}_{ij}$ is the weight connecting the input $x_j$ and the $i$th neuron in the hidden layer. Similarly, $y_l, h_l$ are the output and the local field of the $l$th neuron in the output layer, $w_{li}$ is the weight connecting $\bar{y}_i$ and the $l$th neuron in the output layer, M is the dimension of the input vector, N is the number of neurons in the hidden layer, L is the dimension of output vectors. Here we apply the linear transfer function to the output layer for simplifying the analysis. For function approach, it is applicable directly. For pattern recognition and classification, one can further apply nonlinear transfer functions to assign labels to output vectors after finishing the training.

Figure 1(a) shows the architecture of a GVM, while Figure 1(b) shows that of a typical SVM. We plot two output neurons in the former and a single one in the latter, to emphasize that our method is directly applicable for multi-classification problems while the SVM method is usually for binary decision problems. For practical applications of the SVM method, one usually designs several bi-classification machines to perform the multi classifications. The essential difference from the SVM is that we use general vectors $\overline{w}^\eta \in R^M$ to replace support vectors $x^\eta \in R^M$. To find the solution of the general vectors, we apply the MC algorithm described in next section.

## 3. Monte Carlo algorithm

The MC algorithm is for establishing the correct response on the training set. A three-layer network maps input vectors to output vectors by two steps of transformation. That is: the Hidden layer maps the M-dimension input vectors of samples into N-dimension vectors in the 'feature space', and then the output layer maps them into L-dimensional output vectors. The two layers resemble two coupled mirrors. Simultaneously changing two mirrors, or fixed on one and changing another, could both establish the desired input-output correspondence. The SVM algorithm applies the former strategy. It adjusts the hidden-layer by selected support vectors, $x^\eta \in R^M$, and calculates the output-layer by the linear optimization theory. Introducing the support vectors is the soul of the SVM method. It decreases dramatically the freedom of choosing the hidden-layer parameters, and reduces the solution to be a linear optimization problem in the feature space. This treatment, on the other hand, imposes restriction on the projective vectors since the support vectors can only selected from the input vectors of samples.

Our idea is different. Let us denote the $\eta$th row of the weight matrix $\overline{\mathbf{w}}$ by $\overline{w}^\eta \in R^M$, and call it as the $\eta$th weight vector of the matrix. We give up the support vectors and apply directly the weight vectors to collect the feature of input vectors. To perform the training, we randomly initialize the parameters in the output layer and fixed them afterwards, and then adjust the 'hidden-layer mirror', i.e., the weight vectors $\overline{w}^\eta$ as well as transfer function coefficients and neuron biases, to find the solution. Because there are huge amount of parameters in the hidden layer, the possibility to find solutions with the fixed output layer is still quite high. One can in principle adjusts simultaneously parameters in both layers to find the solution. However, once the 'output-layer mirror' changes, the 'hidden-layer mirror' should adjust accordingly to march the change, which may induce abundant computer time of simulation.

For sake of the simplicity, we set $w_{li} = \pm 1$ randomly to the output layer. The parameters in the hidden layer are also randomly initialized within their available ranges (will be defined in the next section). To supervise the training, a cost function, $\Gamma = \Gamma(y^\mu, t^\mu), \mu = 1, ..., P$ , is constructed by training samples, where $t^\mu$ represents the actual output of the learning machine under the input $x^\mu$. In sections 6 we shall show how to construct $\Gamma$ in detail.

To start the training, we set all of the parameters to random numbers in their available ranges, and calculate the local fields of neurons, $\bar{h}_i^\mu, h_i^\mu, \mu = 1, ..., P; i = 1, ..., N; l = 1, ..., L$, as well as the function $\Gamma$. We then repeatedly apply the following procedure to find hidden-layer parameters: Randomly adapt one hidden-layer parameter to a new value in its available range, and calculate the changes in $\Gamma$; If it does not become worse, accept the adaptation and renew the local fields as well as the outputs of neurons and the cost function $\Gamma$, otherwise give up the adaptation. In more detail, the hidden layer is renewed by the follow rules:

（a）If $\bar{w}_{ij} \to \bar{w}_{ij} + \varepsilon$, then $\bar{h}_i^\mu \to \bar{h}_i^\mu + \varepsilon x_j^\mu, \bar{y}_i^\mu = f_i(\beta_i \bar{h}_i^\mu)$　　(3)

（b）If $\beta_i \to \beta_i + \varepsilon$, then $\bar{h}_i^\mu \to (\beta_i + \varepsilon)\bar{h}_i^\mu, \bar{y}_i^\mu = f_i(\beta_i \bar{h}_i^\mu)$　(4)

（c）If $b_i \to b_i + \varepsilon$, then $\bar{h}_i^\mu \to \bar{h}_i^\mu - \varepsilon, \bar{y}_i^\mu = f_i(\beta_i \bar{h}_i^\mu)$　　(5)

The output layer is renewed by

$$h_l^\mu \to h_l^\mu + w_{li}(\bar{y}_i^\mu - \bar{y}_i^\mu(old)) \qquad (6)$$

where $\bar{y}_i^\mu(old)$ represents the value before adaptation. The renew operations are performed over $\mu = 1,...,P; l = 1,...,L$. For a particular application, one can specify only certain classes of parameters as changeable ones and keeps the others fixed. For designing learning machines with parameters taking continuous values, one can set $\varepsilon$ to be small random numbers. It is not necessary to limit $\varepsilon$ to be sufficiently small. For parameters with discrete states [16-17], the parameter $\varepsilon$ is set to push the parameter jumping from the present state to a neighboring state randomly. The training is stopped until $\Gamma \le \Gamma_0$ or after a sufficient long training time, $t \ge t_0$.

Our algorithm does not need to develop the entire network which needs about $O(NMP + NLP)$ multiply-add operations. Each adaptation in induces only $O(P + LP)$ multiply-add operations and the adaptation accepted is optimum for the whole training set in the statistical sense. The examples shown in the application sections will indicate that the training time of our algorithm is practical for various applications.

Applying the MC algorithm, the goal of extracting the feature of samples is approached by simply projecting the input vectors into the weight vectors in the hidden layer. The projections are considered as the 'features'. Though a single weight vector may extract less information than a support vector does, the unlimited amount of weight vectors can offset this drawback. Indeed, none has proved that support vectors are the best ones. In the case when the training set is quite small, the support vector method has obvious limitation. We suppose that the most optimal vector set that could maximally extract the sample feature should be searched over the entire vector space, and the MC algorithm have such a potential. In principle, more weight vectors mean more feature information. As a consequence, we prefer large learning machine to small ones. The over-training problem induced by large network size can be suppressed by controlling the design risk using strategies introduced in next two sections.

## 4. Controlling the structural risk

In this section we show how to control the structural risk of a GVM by controlling that of single neurons. The response of a function to a small input change can be expressed as

$$\delta y \sim \delta x \frac{\partial \Phi}{\partial x} + \frac{1}{2} (\delta x)^2 \frac{\partial^2 \Phi}{\partial x^2} + \cdots. \qquad (7)$$

Therefore, the moments of derivative of $\Phi(x, \Lambda)$ determine the input-output sensitivity of the function. In application, the second moment is usually applied to define the risk. The second moment of the derivative of the $i$th neuron in the hidden layer to the $j$th and $k$th components of an input vector is $\gamma_{jk}^i \equiv \frac{\partial^2 \overline{y}_i}{\partial x_j \partial x_k}$, which can be derived explicitly:

$$\gamma_{jk}^i = \beta_i^2 \overline{w}_{ji} \overline{w}_{ki} f_i^{''} (\beta_i \overline{h}_i) \qquad (8)$$

All of the components over indexes $j$ and $k$ determines the structural risk of the $i$th neuron. The structural risk of a GVM is the linear combination of the single-neuron risks since we employ linear neurons in the output layer, i.e., for the $l$th output component of a GVM it has $\frac{\partial^2 y_l}{\partial x_j \partial x_k} = \sum_{i=1}^{N} w_{il} \gamma_{jk}^i$.

The structural risk of a GVM is thus determined by totally LM$^2$N terms of second derivatives. One may employ the average amplitude

$$R_S = < \left\| \sum_{i=1}^{N} w_{il} \gamma_{jk}^i \right\| > \qquad (9)$$

to measure the structural risk of a GVM, where <•> represents the average over index pair *(l,j,k)*. This is a hard task for even calculation, let alone to find the solution of the minimum.

Following Eq. (8), the risk of a neuron is determined by the product of $\beta_i^2, \overline{w}_{ji}, \overline{w}_{ki}$ and $f_i^{''}$, and therefore it is controlled by the amplitudes of these parameters. By limiting the parameters in the ranges

$$\beta_i \in [-c_\beta, c_\beta], \quad \overline{w}_{jk} \in [-c_w, c_w], \qquad (10)$$

and choosing neuron transfer functions which has bounded derivative $f_i^{''}$, we can control the amplitude of $\gamma_{jk}^i$. Typical neuron transfer functions suitable for this requirement include

$$f_i(z) = e^{-z^2} \qquad (11)$$

with $f_i^{''} \in [-2, 0.9]$, and

$$f_i(z) = \tanh(z) \qquad (12)$$

with $f_i^{''} \in [-0.8, 0.8]$. The former is called as the Gaussian neuron transfer function and the latter the sigmoid neuron transfer function.

As a linear combination of risks of neurons, the structural risk of a GVM can be controlled by the parameters $c_\beta$ and $c_w$, and are called as the control parameters. Nevertheless, for the zero-input vector and its nearby vectors, it has $f_i^{''} \approx f_i^{''}(0)$ for all of the hidden-layer neurons if $b_i = 0$ following eq. (1), which results in input-vector-value dependent risks. To avoid this situation, the parameter $b_i$ is given a role in controlling the risk. Without loss of the generality, suppose the input vectors are symmetric to the origin. We assign random values to $b_i$ in the range

$$b_i \in [-c_b, c_b] \qquad (13)$$

with $c_b \sim \max \sum \overline{w}_{jk} x_k^\mu$. As $b_i$ and $\overline{w}_{jk}$ take random values, $f_i^{''}$ takes different values even with the zero input vector, and thus the systematical dependence to the input-vector value is greatly reduced. The parameter $c_b$ is thus also applied as a control parameter.

In principle, as a multi-parameter optimization problem, the global optimal solution should be found by searching the control parameter space. Fortunately, as can be seen from eq. (8) that it is the multiple $\beta_i^2 \overline{w}_{ij} \overline{w}_{ik} f_i^{''}$ determines the risk. One thus can fix the control parameter $c_w$ and searches the parameter $c_\beta$ alone to find the solution. Meanwhile, the control parameter $c_\beta$ can also control the range of the variable $f_i$, and control the distribution of $f_i^{''}$. Sometime one may also searching for a more proper $c_b$, but our investigate indicate that it is not sensitive except in the particular situation near the zero input vector.

In SVM method, the polynomial transfer function

$$f_i(\mathrm{z}) = z^n \tag{14}$$

is also constantly employed. Since $f_i^{''}(\mathrm{z}) = n(n-1)z^{n-2}$, the range of the second derivative depends on the input vectors explicitly. It has relatively big amplitude when n is big, and may increase with the increase of the amplitude of input vectors. This transfer function therefore may induce higher structural risk. Therefore, when sometime we have to apply this function to maximum problem-dependent prior knowledge, additional strategy should be employed to suppress the risk.

## 5. The design risk minimization strategy

Extremely minimizing the structural risk may result in the overtraining for function approach. The goal of function approach is to find a learning machine $\Phi(\Lambda, x)$ satisfying $\Phi(\Lambda, x) = g(x)$, where $g(x)$ is the goal function. Minimizing the empirical risk can approach this relation on the sample set. To generalize to the whole interval, the usual way is to decrease the structural risk of $\Phi(\Lambda, x)$ to pursue the smoothness. Let $\int \|d^2\Phi(\Lambda, x)/dx^2\| dx$ and $\int \|d^2 g(x)/dx^2\| dx$ represent the structural risks of the learning machine and the goal function respectively. Usually, the former is much bigger than the latter since the learning machine is a big system with a large set of parameters. Therefore, the minimization can initially decrease the risk of the former to approach that of the latter. However, extreme minimization may make the former smaller than the latter, and thus the overtraining occurs. This is possible since $\Phi(\Lambda, x)$ and $g(x)$ are not identical functions.

We argue that the DRM strategy can avoid the over minimization and lead to the best fitting. To calculate the design risk, we construct n GVMs satisfying $\Gamma \leq \Gamma_0$ with randomly initialized system parameters at a given set of control parameters. These GVMs are trained using the same set of training samples. We simply apply the $\Phi(x, \Lambda)$ of a GVM to be the response function $\Pi(x)$. The squared error

$$E(\Pi) \equiv \sqrt{\frac{1}{n}\sum_{i=1}^{n}(\|\Pi_i(x) - <\Pi(x)>\|)^2} \qquad (15)$$

of the response functions $\Pi_i(x), i = 1,...,n$ of n GVMs defines the design risk, where $<\Pi(x)>$ is the average response function. Each $\Pi_i(x)$ may be considered as a random fluctuation around the goal function $g(x)$, and thus one may expect $<\Pi_i(x)> = g(x)$ for sufficient large n since the random fluctuations may offset with each other. In this case, $E(\Pi)$ equates to the average fitting error defined as

$$<\Theta(\Pi)> \equiv \sqrt{\frac{1}{n}\sum_{i=1}^{n}(\|\Pi_i(x) - g(x)\|)^2} \qquad (16)$$

Therefore, the minimum of the design risk gives that of the average fitting error, and defines the best control parameter set.

For pattern recognition, the moments of derivatives in eq. (7) alone could not determine the input-output sensitivity. In this problem, the outputs of the learning machine should be separated by a separating margin for classifying patterns into different classes. If the $\mu th$ sample is belong to the $\nu th$ class, the learning machine should be response as $y_\nu^\mu - y_l^\mu \equiv 2d > 0$ for $l \neq \nu$. A variation of the $\mu th$ sample, $x = x^\mu + \delta x$, is expected to be classified into the same class that the $\mu th$ sample belongs. Inputting the variation into the learning machine, one have $y = y^\mu + \delta y$. The variation can be correctly classified into the $\nu th$ class if only $y_\nu - y_l > 0$ for $l \neq \nu$. This condition can be represented as $2d + \delta y_\nu - \delta y_l > 0$ alternatively. Thus, the correct classification is determined not only by the amplitudes of derivatives of $\Phi(\Lambda, x)$ which affects the amplitudes of $\delta y_\nu$ and $\delta y_l$, but also by the separating margin. The input-output sensitivity in this case is the flexibility of the condition $2d + \delta y_\nu - \delta y_l > 0$ being violated with input variations of $x = x^\mu + \delta x$. Therefore, the input-output sensitivity here is not equivalent to the mathematical concept of the structural risk. To decrease the sensitivity, one may fix the separation margin and decrease the amplitudes of derivatives of $\Phi(\Lambda, x)$, or fix the latters and increase the former. For this reason, the width of the separating margin $d$ is also applied as a control parameter in our algorithm.

We argue that the design risk is also the favorable indicator for supervising the learning machine for pattern recognition. We denote the correct rate of a GVM on the test set to be $\Pi$ and apply it as the response function to measure the design risk. Designing a set of GVMs to obtain a response function series $\Pi_i, i = 1,...,n$ at a given control parameter set, we obtain two performance indexes, the average correct rate $< \Pi >$ and the dispersion degree of the correct rates. The latter defines just the design risk. We minimize the design risk to pursue a better performance based on following reasons. Firstly, minimizing the design risk is a necessary demand for a design method. If learning machines designed by different users following the same algorithm have big dispersion in recognizing the same test set, the method will be lack of reliability. Secondly, minimizing the design risk can suppress the input-output sensitivity of the learning machine, and can maximally incorporate the common prior knowledge of 'variations of a knowing pattern should be assigned to the same class'. Bigger degree of the input-output sensitivity may induce bigger output fluctuations of $\delta y$, and in turn induce bigger dispersion in the design risk. As a result, minimizing the design risk can usually maximize the average correct rate.

This occurs when real patterns can be considered as the random variations of training samples. In more detail, if the variations of the $\mu th$ pattern can be represented by $x_i = x_i^\mu + \zeta_i$, $E\zeta_i = 0, E\zeta_i^2 = \sigma^2$, then minimizing the input-output sensitivity implies variations with as big as possible mean square $\sigma$ could be classified into the class that the $\mu th$ pattern belongs. It is just this case marches exactly the common prior knowledge, since 'variations of a knowing pattern should be assigned to the same class' put no particular restriction to the variation. Grasping this point may help us to understand the applicable scope of so-called design principles, such as the Occam razor principle, the SRM principle and the DRM strategy in this paper. A principle is for maximizing the common prior knowledge and is thus applicable generally. However, it cannot be expected to always give the optimal solution for particular problems since it may loss the problem-dependent prior knowledge.

## 6. Incorporating the problem-dependent prior knowledge

For function approach, the problem-dependent prior knowledge is about the goal function inducing the data set. In this case, choosing a proper neuron transfer function may better march the feature of the data. For example, if the data come from a polynomial goal function, applying a polynomial transfer function shall achieve a better fitting than using a sigmoid one. Our application examples will show that the more proper neuron transfer function can be selected also by checking the design risk, i.e., a smaller design risk will become smaller when a better neuron transfer function is applied.

For pattern recognition, there are many types of problem-dependent prior knowledge, such as the interpretation of input vectors, the rotational and translational invariance of patterns, the special geometric symmetry of patterns etc. Applying individualized strategies to maximally utilize the problem-dependent prior knowledge is essential for further improve the generalization ability.

The physical interpretation of input vectors may involve the problem-dependent prior knowledge. When a variation $x = x^\mu + \delta x$ of the $\mu$th sample is input to a GVM, deviations $\bar{h}_i \rightarrow \bar{h}_i{}^\mu + \delta \bar{h}_i$ and $h_l \rightarrow h_l{}^\mu + \delta h_l$ in local fields in hidden layer and in the output layer will be induced in turn. In medical diagnosis, for example, each component of an input vector describes a biochemical indicator. As in the Wisconsin breast cancer database [14], the components are endowed the meanings that the more low the value the more normal, and the more high the more likely malignant. As a result, negative $\delta \overline{h_i}$ represents the normal, and positive $\delta \overline{h_i}$ indicates the divergence from the normal. To match this feature, the sigmoid transfer function is preferable than the Gauss transfer function. In other examples as for pattern recognition, an input vector encodes a two-dimensional pattern. A component of such a vector has a standard reference value for a specific pattern. The corresponding component of a new pattern with either great or less value both represent the deviation from the reference value. In this case, the Gaussian neuron transfer function may march the feature better.

Similarly, the property of $h_l^\mu$ should also march the symmetry feature of samples. We introduce several cost functions to control the distribution of local fields of neurons in the output layer. The first one is defined by

$$F_1 = \frac{1}{PL} \sum_{\mu=1}^{P} \sum_{l=1, h_l^\mu s_l^\mu < d}^{L} (h_l^\mu s_l^\mu - d)^2 , \qquad (17)$$

where $s_\nu^\mu = 1$ and $s_l^\mu = -1$ for $l \neq \nu$ if the sample belong to the $\nu th$ class. When it is minimized to $F_1 = 0$, the local fields of neurons in the output layer satisfy $h_l^\mu s_l^\mu > d$ for all training samples. In this case, it has $h_\nu^\mu > d$ and $h_l^\mu < d, l \neq \nu$ otherwise. The distribution of the local fields is illustrated schematically in Fig. 2(a). We call this function the steep-margin cost function. The second one is

$$F_2 = \frac{1}{PL} \sum_{\mu=1}^{P} \sum_{l=1}^{L} (h_l^\mu s_l^\mu - d)^2 \qquad (18)$$

which compresses $h_l^\mu s_l^\mu$ around $d$. The resulted distribution of $h_l^\mu$ is illustrated in Fig. 2(b). We call it the Gauss-margin cost function.

In certain case, $F_2$ may not be approachable due to the huge amount of training samples. In this situation, introducing the following cost function,

$$F_3 = \frac{1}{PL} \sum_{\mu=1}^{P} \begin{cases} \displaystyle\sum_{l=1, h_l^\mu s_l^\mu < d_1}^{L} (h_l^\mu s_l^\mu - d_1)^2 \\ \displaystyle\sum_{l=1, h_l^\mu s_l^\mu > d_2}^{L} (h_l^\mu s_l^\mu - d_1)^2 \end{cases} \qquad (19)$$

may be helpful. The minimization of it drives $h_l^\mu s_l^\mu$ into the interval $[d_1, d_2]$. When $d_2 >> d_1$ it approaches $F_1$, while when $d_2 \sim d_1$ it approaches $F_2$.

These cost functions can also be presented in another way. For example, the second cost function can be modified as

$$\overline{F}_2 = \frac{1}{PL} \sum_{\mu=1}^{P} \sum_{l=1, l \neq \nu}^{L} (h_\nu^\mu - h_l^\mu - d)^2 \qquad (20)$$

By minimizing this function, $h_\nu^\mu$ and $h_l^\mu$ for $l \neq \nu$ are separated by a distance $d$ for each sample patterns, but the local fields need not to distribute around the origin.

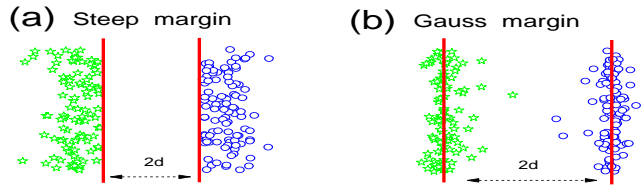

Figure 2. Schematic diagram of separating margin. (a) Steep margin and (b) Gauss margin defined by the cost functions $F_1$ and $F_2$ respectively.

Properly pretreatment of input vectors is also an effective manner to incorporating the problem-dependent knowledge of transform symmetry. For handwritten digit recognition, the patterns have symmetries under small spatial shifts, small rotations, as well as small distortions. For man-made objectives, spatial shifts and angle variations are important but distortions may have no corresponding. For these particular problems, generalization based on random variations is extravagant. To incorporate the particular symmetric restriction, one can construct spurious training samples by the shift, rotation, distortion, tangent distance technique, etc.

[18], and applies them also in the training. The drawback is that big amount of additional calculations is induced for training the machine. Another way is to encode the symmetry property into the input vectors of samples. The so-called Gradient-base feature-extraction algorithm *e-grg* [19], with feature vector encoding eight direction-specific $5 \times 5$ gradient images, is one of the top-performing algorithms for this purpose.

In these cases, especially when particular geometric restrictions are involved, however, the control parameter set with maximum average correct rate and that with minimum design risk may be inconsistent with each other, and the average correct rate on the test set should also be applied as a performance index to identify the best control parameter set.

## 7. Using a joint learning machine

There is a way to further decrease the risk, i.e., combining a huge number of GVMs designed at the same control parameter set to construct a joint learning machine, a J-GVM. The structural risk of a J-GVM is the algebraic average of the GVMs. With the small risk, one may expect that the J-GVM has better performance than a single GVM.

The GVMs composing the J-GVM are statistically identical, since there are obtained with the same training set at the same control parameter set. For an individual GVM, its output involves not only the information related to training samples but also random noise. The output of a J-GVM is an ensemble average of the vast GVMs, the noise part is thus be suppressed. Based on this consideration, the J-GVM needs not necessarily to be constructed using the GVMs at the best control-parameter set of individual GVMs. For pattern recognition, application examples will show that the J-GVM constructed at a control parameter set with GVMs having a relatively big degree of risk may have better performance. In such a case, weight vectors have big freedom to extract the feature of input vectors, and therefore much more features of samples may be extracted. Because of this reason, we ask a sufficient amount of GVMs to construct the J-GVM so as to offset the training-sample independent fluctuations. This strategy is similar to the conventional ensemble methods [6,11].

# 8  Application for function approach and smoothing

We apply a $M - N - 1$ GVM for function approach and smoothing. The MC algorithm is applied to train a GVM in response to samples, which effect is measured by the empirical risk

$$F_e = \sqrt{\frac{1}{P} \sum_{\mu=1}^{P} (t^\mu - y^\mu)^2} \qquad (21),$$

where $t^\mu$ and $y^\mu$ are the target and actual output of the $\mu$ th sample. The training is stopped when the empirical risk is smaller than a threshold or a maximum number of MC steps approached. The latter stop condition is for smoothing noise data sets, in which case the former condition may not be approached.

Our examples include fitting the sin function $g(x) = \sin(x)$, the sinc function $g(x) = \sin(x)/x$, the Hermit 5th polynomial $g(x) = (63x^5 - 70x^3 + 15x)/8$, the Hermit 7th polynomial $g(x) = (429x^7 - 693x^5 + 315x^3 - 35x)/16$, and the 2-D sinc function $g(x, y) = \sin(\sqrt{x^2 + y^2})/\sqrt{x^2 + y^2}$. For data smoothing, we add the white noise to the sinc function as

$$g(x) = \frac{\sin(x)}{x} + \zeta_i, E\zeta_i = 0, E\zeta_i^2 = \sigma^2. \qquad (22)$$

## 8.1 Finding the best control-parameter set

We prepare three noise-free training sets from goal functions of the sin, sinc and Hermit 5th polynomial for function approach. Each set has

20 uniformly distributed samples $[x_i, g(x_i)]$ from interval of $x_i \in [-c_x, c_x]$. The set of the sin function is from one period of $c_x = \pi$. The sinc function is in the interval of $c_x = 10$. The domain of the Hermit polynomial is *[-1,1]* with $c_x = 1$. The stop condition is $F_e < 10^{-4}$ for all the three sets. One noise set from the sinc function, i.e., 100 samples with $\sigma = 0.1$ as in the reference [4], is applied for data smoothing. The training is stopped for this set after $10^5 N$ MC steps. The Gauss transfer function is adopted for all training sets. The parameters of the GVM are limited in the intervals $\beta_i \in [-c_\beta, c_\beta]$ , $\overline{w}_i \in [-10/c_x, 10/c_x]$ , $b_i \in [-10, 10]$ , correspondingly. The factor $1/c_x$ results $\max[\overline{w}_i x] = c_b = 10$, which guarantees the optimal value of $c_\beta$ keep universal roughly for different samples (see examples below). With these settings, only $c_\beta$ remains to be the control parameter. At each point of $c_\beta$, 500 GVMs are designed with random initializations in the specified ranges of system parameters. They are applied to calculate the average structural risk, average fitting error and design risk.

Because there is only one neuron in the input and output layers respectively, the structural risk defined by eq. (9) can be calculated directly. The first row of Fig. 3 shows $<R_s> - R_s^g$, and the second row shows the $<\Theta(\Pi)>$ and $E[\Pi]$, as functions of the parameter $c_\beta$ for the four sample sets correspondingly. The average structural risk $<R_s>$ is calculated over the risk $R_s$ of GVMs. The structural risk $R_s^g \equiv \| d^2 g(x) / dx^2 \|$ of the goal function is a constant and is applied as a reference line for $<R_s>$. It can be seen that $<R_s>$ decreases rapidly with the decrease of $c_\beta$. This fact indicates that the risk $<R_s>$ do can be decreased by decreasing that of individual neurons. With the further decrease of $c_\beta$ , $<R_s>$ shows a minimum. However, it is usually inconsistent to that of $<\Theta(\Pi)>$. For the first and last training sets, the differences between the minima of $<R_s>$ and $<\Theta(\Pi)>$ are slight, while for the second and third sets, the differences are remarkable. Particularly, for the third set, the minimum of $<R_s>$ is around $c_\beta = 1.0$ at which $<\Theta(\Pi)> = 1.4 \times 10^{-2}$, while the minimum of $<\Theta(\Pi)>$ appears at $c_\beta = 0.48$ with $<\Theta(\Pi)> = 2.9 \times 10^{-3}$. Therefore, minimizing the structural risk of a learning machine does not necessarily converge to the best machine.

It can be realized from the figure that the difference is induced by the over minimization, i.e., the risk of the learning machine is minimized to a value that is smaller than that of the goal function. In the cases of Fig. 3(b) and 3(c), $<R_s> - R_s^g$ shows negative intervals and the difference between the minima of $<R_s>$ and $<\Theta(\Pi)>$ are remarkable. In the cases of Fig. 3(a) and 3(d), $<R_s> - R_s^g$ keeps non-negative, and the minima of $<R_s>$ are close to those of $<\Theta(\Pi)>$ correspondingly. Nevertheless, because $R_s^g$ is unknown for practical applications, one cannot judge whether the over minimization occurs, and thus can not apply the structural risk criterion with confidence to judge whether the best fitting is achieved. This effect cannot be generally avoided since $\Phi(x, \Lambda)$ and $g(x)$ are different functions in principle. Minimizing the risk of the former cannot necessarily converge to that of the latter.

On the contrary, the minima of $E[\Pi]$ are consistent approximately to those of $<\Theta(\Pi)>$ for either data set. In the cases of the first two sets, $E[\Pi]$ and $<\Theta(\Pi)>$ coincide almost with each other, indicating the satisfying of $<\Pi_i(x)> = g(x)$. For the last two sets, $E[\Pi]$ and $<\Theta(\Pi)>$ show the similar dependence on $c_\beta$, indicating that $<\Pi_i(x)>$ can still give the best approach to $g(x)$. Therefore, the minimum of the design risk defines the best control-parameter set for function approach and smoothing.

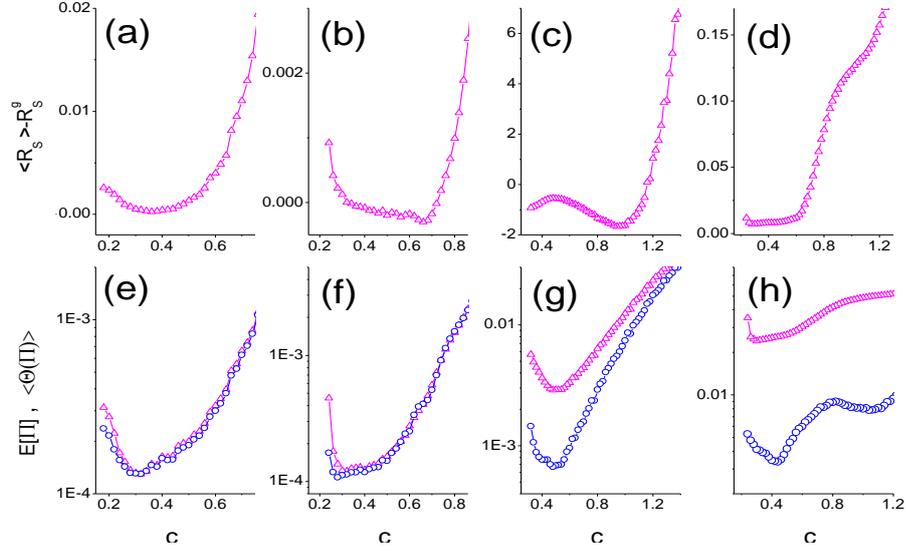

Figure 3. The first row shows $<R_S>-R_S^g$, and the second row shows $<\Theta(\Pi)>$ (up-triangles) and $E[\Pi]$ (circles) as functions of $c_\beta$. (a) and (e): sin function; (b) and (f): sinc function; (c) and (g): hermit polynomial; (d) and (h): sinc function with noise amplitude $\sigma = 0.1$.

The inconsistence between $E[\Pi]$ and $<\Theta(\Pi)>$ can be understood by investigating the fitting process of the noised data. At the control parameter set with the minimum design risk, Fig. 4(a) shows two fitting curves of different GVMs for a set of noise data with $\sigma = 0.2$ created by eq. (22). These curves converge to almost the same function, but have systematic deviation from the goal function. Obviously, this is not the problem of the DRM strategy, instead, it is an effect of finite training samples. For a finite training set, the noise may induce the systematic divergence from the goal function. Suppose we have two training sets and each has just several sample points for example, then we cannot distinct which one is the noised data and which one from the goal function in principle without additional prior knowledge. The noised data set itself also defines a 'goal function'. Applying the DRM strategy approach this goal function and results the systematic deviation from the real goal function. To decrease the deviation induced by this mechanism, one has to applying small-noise-amplitude samples or getting more training samples.

This effect may also exist for noise-free finite training samples. Giving just several points, there is big uncertainty to judge which goal function they come from. Applying more training samples can decrease the uncertainty. Furthermore, different from the noised data, the deviation may be dramatically suppressed by choosing more proper neuron transfer function which marches the real goal function better, as well be shown in section 8.6.

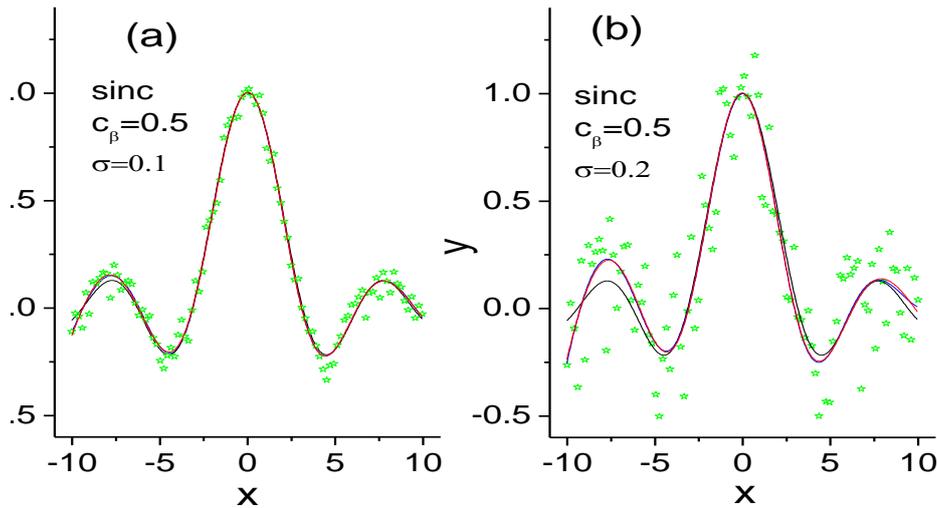

Figure 4. Function smoothing for samples with noise intensity of $\sigma = 0.1$ (a) and $\sigma = 0.2$ (b) . 100 samples (stars) are distributed uniformly in the interval of [-10,10]. The goal function is the 1D sinc function (black line). In the plot, two fitting curves are shown, which are almost overlapped with each other.

## 8.2 Improving the fitting by increasing the learning machine size

As can be seen from Table I, increasing samples can greatly increase the fitting precision. This is generally true for either our algorithm or conventional methods. Here we emphasize that we can also improve the fitting precision by increasing hidden-layer neurons. Increasing the hidden-layer neurons from 100 to 1000, the fitting precision may be further improved by up to threefold. This is a remarkable difference to conventional methods. The BP method follows the 'Occam razor principle' to pursue neural networks with as smaller as possible size. If a hidden-layer with 10-100 times of samples is applied then serious over-fitting must be arisen. For the SVM method, the hidden-layer neurons are limited by the amount of the samples. The hidden-layer neurons could not exceed the number of samples.

Table I : Fitting precision vs. sample amount and machine size

| Samples | 10 | | | 20 | | |
|---|---|---|---|---|---|---|
| Data | Sin | sinc | hermit | sin | sinc | hermit |
| GVM(1-100-1) | $4.5 \times 10^{-4}$ | $4.5 \times 10^{-3}$ | $2.5 \times 10^{-2}$ | $1.3 \times 10^{-4}$ | $1.2 \times 10^{-4}$ | $2.9 \times 10^{-3}$ |
| GVM(1-1000-1) | $1.3 \times 10^{-4}$ | $2.6 \times 10^{-3}$ | $1.0 \times 10^{-2}$ | $1.2 \times 10^{-4}$ | $1.1 \times 10^{-4}$ | $1.0 \times 10^{-3}$ |
| J-GVM(1-100-1) | $5.3 \times 10^{-5}$ | $2.4 \times 10^{-3}$ | $2.5 \times 10^{-2}$ | $1.0 \times 10^{-5}$ | $3.6 \times 10^{-5}$ | $2.9 \times 10^{-3}$ |
| Spline | $3.2 \times 10^{-4}$ | $1.5 \times 10^{-2}$ | $8.2 \times 10^{-2}$ | $1.5 \times 10^{-5}$ | $1.1 \times 10^{-3}$ | $1.3 \times 10^{-2}$ |

## 8.3 Fitting by a J-GVM

To obtain GVMs, the training is stopped when $F_e \leq 10^{-4}$. Therefore, the fitting precision of a GVM could not beyond this threshold. Table I shows that applying a J-GVM can usually obtain better result than a GVM. The J-GVM is constructed by the 500 GVMs trained at the best control parameter $c_\beta$ where $E[\Pi]$ takes the minimum. For certain data sets the precision can be improved by even one order, which is much higher than the empirical risk.

## 8.4 Comparison to conventional algorithms of function approach

Table I also shows the corresponding results using the widely applied spline algorithm for function approach. For the last two sets, GVMs get obviously better performance. For the data of sin function, the spline algorithm achieves the same precision. However, this is because we apply the training stop condition of $F_e \leq 10^{-4}$, which limits the precision of the fitting. If one changes the stop condition to $F_e \leq 10^{-5}$, one can further improve the precision of GVMs to the order of $10^{-5}$. The J-GVM is constantly better than the traditional algorithm for either data set.

As to compare to the SVM, we would like to mention the example shown in text book of Vapnik [4]. In this example, 100 samples for the sinc function are applied as training set. When choosing 14 samples from this set to be support vectors, the fitting curve already shows big diverges over a amplitude of 0.1 from the goal function. In our case, even for the training set with only 10 samples, one can approach a precision of $< \Theta(\Pi) >= 4.5 \times 10^{-3}$ using a GVM. The fitting curve is already indistinguishable visually from the goal function.

## 8.5 Training time

For the sake of comparison, the training time is measured by the CPU time of commonly used personal computer (specifically 2.0 GHz). Figure 5 shows the average training time of a GVM as a function of $c_\beta$ in the case of the sinc goal function with 20 samples. It can be seen that the training time increases rapidly with the decrease of $c_\beta$. The training time may increase exponentially when the GVM becomes too small. We have checked that for $N \leq 20$, the training fails for the condition of $F_e < 10^{-4}$ cannot being approached within a reasonable training time. On the contrary, the training time decreases slightly with the increase of the machine size. Together with the fact that large machines may improve the fit precision, our algorithm thus prefers large machines than small ones.

Over-fitting problem of large machines can be suppressed by further decreasing the design risk.

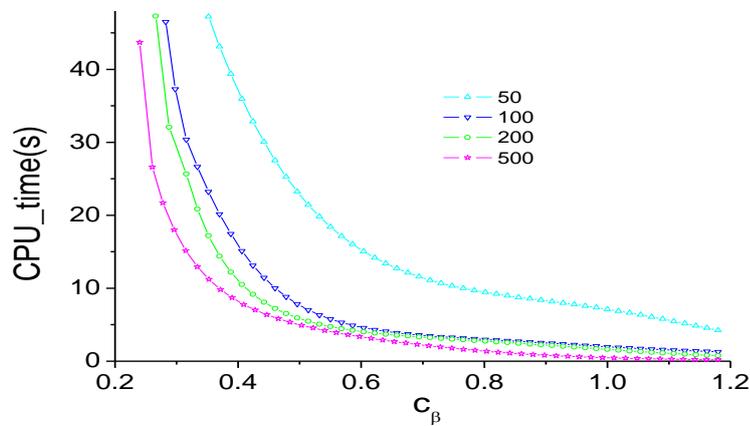

Figure 5. Training time as functions of the machine size.

## 8.6 Improving the fitting by proper neural transfer functions

The BP method is derived with the sigmoid neuron transfer function and thus has no choice for the transfer function. Choosing the transfer function (kernel function) remains a vexing issue in SVM method. One usually needs to search for different function for different problem. As for fitting the sinc function, a complex form $f_i(x, x^i) = 1 + x^i x + (1/2)|x - x^i|(x \wedge x^i)^2| + (1/3)(x \wedge x^i)^2$ is employed, where $x^i$ represents a support vector [4]. Our algorithm does not sensitively depend on the transfer function, as shown by above examples that favorable fittings are obtained with the Gauss transfer function for either data set.

However, when the transfer function marches the feature of the goal function better, one may obtain better result. Figure 6 shows the fitting results using the Gauss, sigmoid, and polynomial neural transfer functions for the data sets of the goal functions sin, sinc and Hermit 5th polynomial. In each set there are 10 sample points. The polynomial transfer function is defined by a six order polynomial $f(h) = h^6$.

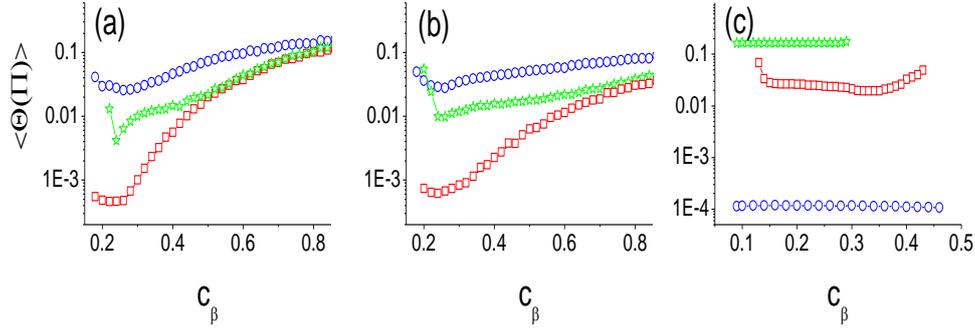

Figure 6. The fitting precision as a function of $c_\beta$ by use of the Gauss (a), Sigmoid (b), and polynomial (c) neural transfer functions. The squares, stars, and circles are for data sets from the sin, sinc, and Hermit polynomial goal functions, correspondingly.

It can be seen that fitting with the Gauss transfer function can approach better results for all the three data sets than with the sigmoid transfer function, but the difference is not remarkable. By using the polynomial transfer function, the difference is big. For the data set of the sinc function, the best precision is about $<\Theta(\Pi)>=0.16$. Indeed, even the empirical risk can only be minimized below to $F_e = 0.10$ for this set. For the data set of the sin function, a precision of $<\Theta(\Pi)>=0.02$ can be approached, but is still quit worse than those using another two transfer functions. For the data set of the Hermit polynomial function, however, a much high precision is achieved. With the training stop condition of $F_e \leq 10^{-4}$, the fitting precision remains below $<\Theta(\Pi)>=1.1\times 10^{-4}$ as Fig. 6(c) indicated. With $F_e \leq 10^{-6}$, one can indeed achieve $<\Theta(\Pi)>=1.1\times 10^{-6}$. These facts imply that the goal function can be recovered with remarkable precision. We have checked that the high-precision fitting can always achieved for this data set if applying a transfer function of $f(h) = h^n$ with $n \geq 5$. The perfect precision obviously comes from the fact that the polynomial transfer function marches the feature of this data set well. Similar discussion is applicable for explaining the results in Fig. 6(a) and 6(b), where the Gauss transfer function marchers the data sets better than the sigmoid one. The more favorable neural transfer function can be chosen following the design risk criterion or the prior knowledge of the data source. When a more favorable neural transfer function is applied, the design risk will become more small.

## 8.7 Applying as a universal fitting machine

For practical applications, one usually needs not necessarily to search for the best-control parameter set. It can be realized from the above examples that the best fitting achieves around $c_\beta = 0.5$ for all of the data sets. Applying a sufficient large machine with Gaussian neuron transfer function, we can perform the function approach for various data sets by fixing $c_\beta$ at $c_\beta = 0.5$. Figure 7 shows that applies a 1-200-1 GVM with Gaussian transfer function we fulfill the fitting well for several complex goal functions. The first data set comes from the sinc goal function in the interval of [-20,20], and the second from the sin goal function in the interval of $[-5\pi,5\pi]$. The third set is from the Hermit 7th polynomial. The last set is from a square wave in [-10,10]. For each set, only 20 samples are applied. For more complex goal functions, the training may be not achieved (the empirical risk cannot be decrease to the target threshold of $F_e$) with $c_\beta = 0.5$. In this case, one can allow the computer to increase the hidden-layer neurons until the training is achieved. Therefore, for usually application of the function approach, the calculation of $E[\Pi]$ using a large number of GVMs can be avoided. This kind of calculation is necessary only when the higher fitting precision is essential.

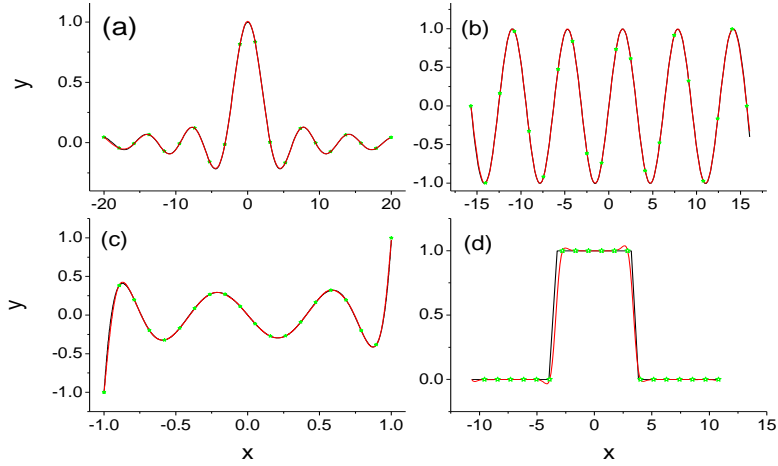

Figure 7. Fitting different data set using a *1-200-1* GVM with the Gaussian transfer function. (a) The sinc function in [-20,20]. (b) The sin function in $[-5\pi, 5\pi]$. (c) The hermit 7th polynomial in [-1,1]. (d) A piecewise step function in [-10,10]. There are 20 samples in either sets.

The algorithm can also be applied directly to high dimensional fitting. The SVM method gives a desirable fitting precision by chosen 153 support vectors from $20 \times 20$ samples [4] of the two-Dimensional sinc function. Applying our strategy to design a 2-1000-1 GVM, Fig. 8 shows that better result is achieved by use of $10 \times 10$ sample points only.

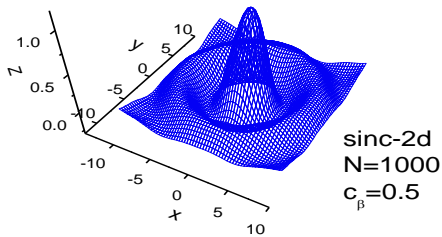

Figure 8. Fitting 2D sinc function using 10x10 samples with a *2-1000-1* GVM.

## 8.8 The role of bias

Here we provide examples to explain the role of the neuron bias. The dependence of the structural risk on the control parameter $c_b$ is somewhat complex. This parameter involves implicitly in $f_i^{''}$. Particularly, if set $c_b = 0$, it has $f_i^{''}(0) = 0$ for each neuron with input $x = 0$ in applying above transfer functions, i.e., it lead to the smallest structural risk with the zero-vector input. As a consequence, it results the response of the GVM rigid to inputs around the zero-vector input and thus the fitting may become quite difficult. Figure 9 shows two examples of such an effect, where tow GVMs with the Gaussian transfer function designed at $b_i = 0$ are applied to fitting the 1D sinc function and the sin function. Around the origin, one can clearly see big deviations from the goal functions, indicating that at the origin the fitting curve is stiff and thus difficult to be approached.

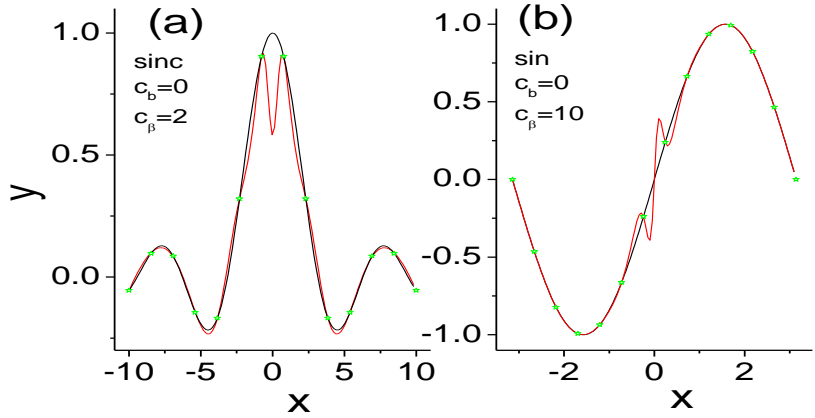

Figure 9. The fitting with $b_i = 0$ for the sinc function (a) and the sin function (b).

To disentangle this problem, our idea is to decouple the dependence of $f_i^{''}$ on the specific value of inputs, and so as to guarantee the sensitivity of a GVM been determined by the intrinsic behavior of the data set instead of values of input vectors. For this purpose, we set $c_b \sim \max[\bar{w}_{jk} x]$, in which case the random initializations of $\bar{w}_{jk}$ and $b_i$ led to the distribution of $f_i^{''}$ insensitively depend on the specific value of input vectors.

## 9 Pattern Recognition

We perform a standard handwritten digit recognition task to show how to design the GVM for pattern recognition. The dataset MNIST [13]

has 60000 training samples and 10000 test samples. Each sample is represented by a $28 \times 28$ dimensional vector. To perform this task, the BP method trains a $28 \times 28 - N - 10$ multi-classifier[13], while the SVM method usually designs ten binary-classifiers.

The GVM has the $28 \times 28 - N - 10$ architecture, too. The input vector $x^\mu$ of the $\mu$th sample represents the $\mu$th pattern. The original data $x_i^\mu$ takes integer of $(0,255)$. We rescale the component $x_i^\mu$ by $0.1*(x_i^\mu - 100)$ as input variable. The output target vector $y^\mu$ encodes the digit $\nu$, $\nu \in (0,...,9)$, which is defined by

$$y_l^\mu = \begin{cases} > 0, l = \nu + 1, \\ \leq 0, otherwise. \end{cases}$$

Such a GVM responses correctly to the training set if only $h_l^\mu s_l^\mu > 0$, for $\mu = 1,...,P; l = 1,...,L$, where $s_l^\mu = sign(y^\mu)$. In this case, the empirical risk vanishes. For sake of simplicity, we fix $c_w = 1$, $c_b = 100$ and $w_{il} = \pm 1$ below.

## 9.1 Finding the best control-parameter set

To decrease the input-output sensitivity, we can decrease the control parameter $c_\beta$ at fixed $d$, or increase $d$ at fixed $c_\beta$. In this way, we can search the best control-parameter set alone only one parameter axis. At a control parameter set, n GVMs are designed to calculate the average recognition rate $<\Pi>$, the average structural risk $<R_s>$, and the design risk $E[\Pi]$. The design risk is calculated by using the recognition rate $\Pi$ of a GVM on the test set to be the response function. To estimate the structural risk, only the second derivatives $\gamma_{jj}^i$ of the hidden-layer neurons are involved in eq. (9), otherwise the calculation should be a hard task. We have checked by full calculation of $\gamma_{jk}^i$ on a small training set and found no qualitative difference.

We first show the dependence of $<\Pi>$, $<R_S>$ and $E[\Pi]$ on $c_\beta$ with $d$ fixed. We train 500 GVMs with the first 1% MNIST training samples by applying $F_2$ with $F_2 < 1$ to be the cost function and the Gaussian transfer function to be the neuron transfer function. The results are shown in Fig. 10. The circles and triangles in Fig. 10 show the results with $N=1000$ and $N=3000$ at $d=30$ and $d=100$ respectively. It can be seen that, $<R_S>$ and $E[\Pi]$ decrease monotonously with the decrease of $c_\beta$. The recognition rate $<\Pi>$ increases rapidly with the decrease of $c_\beta$ initially. After the turning point of $c_\beta \approx 0.005$, it turns to decrease.

Therefore, the best control parameter set is not at the minimum neither of the structural risk nor the design risk. In this case, we have to employ also the average recognition rate as a performance indicator variable. The best control-parameter set is determined by combining the design risk and the average recognition rate, which is a balance between a high recognition rate and an acceptable design risk. In this example, the turning point of the average rate can define the best control parameter set since at which the design risk is acceptably low.

Figure 10 reveals more. Firstly, applying large machines can not only increase the recognition rate but also decrease the design risk. At the turning point, the average recognition rate is 88.8% for N=1000 and 89.5% for N=3000, while the design risk is about 0.2% for N=1000 and about 0.1% for N=3000. Secondly, it reveals that the turning point on $c_\beta$ is insensitive to the machine size. Either for N=1000 and N=3000, it appears around $c_\beta = 0.005$. We can thus fix the parameter $c_\beta$ to this value in our following studies.

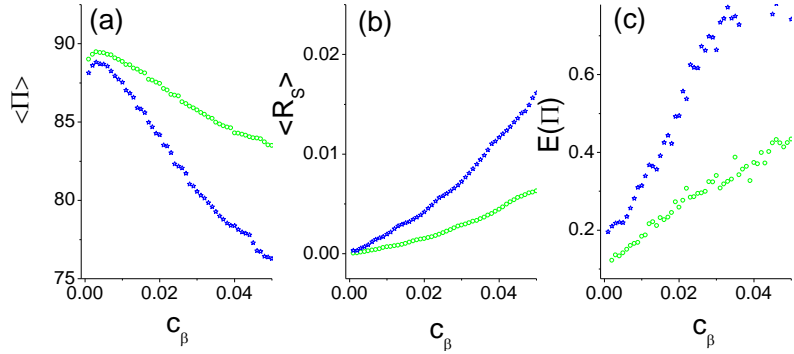

Figure 10. The average recognition rate (a), the structural risk (b), and the design risk (c) as functions of control parameter $c_\beta$. The stars for $d = 30$ and $N = 1000$, and the circles for $d = 100$ and $N = 3000$, respectively.

We then study the dependence of $< \Pi >$, $< R_s >$ and $E[\Pi]$ on $d$ at $c_\beta = 0.005$. Fig. 11 shows the results for *N=3000*. It can be seen that $< \Pi >$ increases rapidly with the increase of $d$ initially, and becomes decrease after the turning point around *d=120*. The design risk decreases monotonously with the increase of $d$. Thus, over maximizing the separating margin may also result the overtraining for this data set. The structural risk $< R_s >$ indeed increases slowly. This is because, though the parameter $c_\beta$ is fixed, the MC adaption may induce $\beta_i$ concentrating slightly towards the boundaries of the specified interval and thus increases the structural risk slightly.

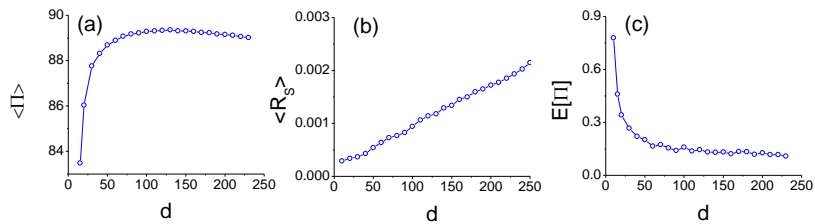

Figure 11. The average recognition rate (a) the structural risk (b) and the design risk (c) as functions of control parameter $d$ at $c_\beta = 0.005$ and $N = 3000$.

In principle, the large the separating margin, the big the probability that a variation of a sample being classified into the same class. This is, however, true when the test set can be considered to be random variations of training simples. Here we illustrate this guess by examples. We construct following two test sets using virtual samples. The first one is created by adding random noise to input vectors of the first 1% samples as $x_i = x_i^\mu + \zeta_i$, $E\zeta_i = 0, E\zeta_i^2 = \sigma^2$ with $\sigma = 80$. For each sample, 10 virtual samples are created and thus totally 6000 samples are involved in this set. The second one is obtained by shifting each of the 1% sample patterns with 2 pixel units to adjacent positions, which gives totally 4800 samples then. Figure 12 show the average recognition rate measured on these tow test sets for GVMs with N=3000 as a function of the control parameter *d*. For comparison purpose, the result for the original test set is also shown. The other control parameters keep same as in the Fig. 11. One can see that with the increasing of the separating margin, $<\Pi>$ for the noise set increases monotonously, while for another two sets the over-training effect appears after the same turning point.

These results indicate that the maximum-margin strategy applied by the SVM method is correct quantitatively for random variations. In other ward, it is correct generally for maximizing the common prior knowledge of 'variations of a knowing pattern should be assigned to the same class' without adding particular restriction on the variations. For practical applications, as for handwritten digits, patterns cannot be considered to be complete random variations of the training samples since they are restricted by the particular geometry of digits. Therefore, the separating margin may not be applied as the quantitative criterion of best learning machine generally.

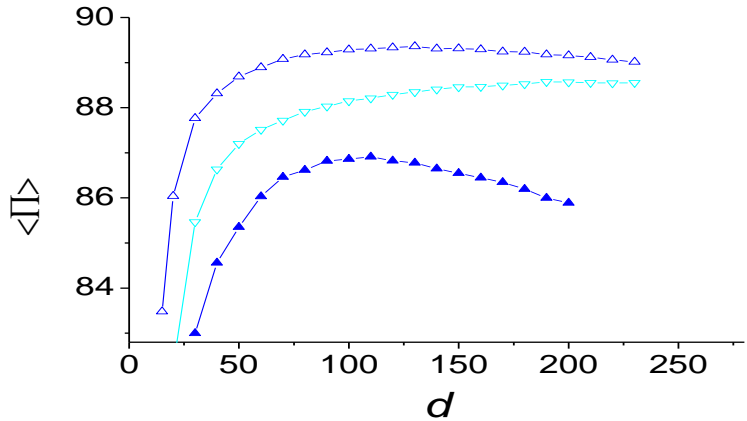

Figure 12. The average recognition rate as a function of $d$ at $c_{\beta} = 0.005$ and $N = 3000$ for the original test set (up-triangles), the test set of random variations (down-triangles), and the test set of shifted samples (solid triangles).

The dependence of $<\Pi>$ on $d$ for the test set of shifted virtual samples is similar to that of using the original test set. This fact can be interpreted as that the shift operation keeps the geometry of digit patterns. It also implies that the particular geometry determines the turning point. Because it gives the same turning point as using the original test set, one can apply this set to find the best control parameter set and apply the original test set also to train the learning machine, in which way the samples may be maximally utilized.

## 9.2 Improving the recognition rate by increasing the machine size

Increasing the machine size can extract more features of simples, and thus can increase the generalization ability. Figure 13 shows the dependence of the average recognition rate on the machine size, which indicates that increasing the size can improve the recognition rate monotonously, though the effect may become saturated when the size is large enough. The reason is that each weight vector extracts information of samples from a different angle, and thus the more the weight vectors, the more features of samples can be extracted.

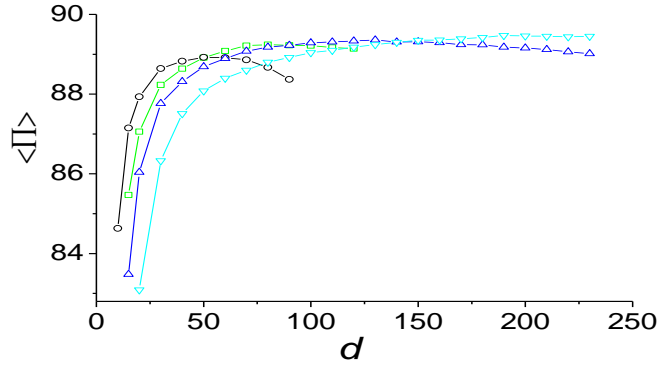

Figure 13. The dependence of the average recognition rate on the machine size. The hollow circles, squares, up-triangles and down-triangles are for GVMs with *N=1000、2000、3000、6000*, correspondingly.

The figure shows that the best control parameter set depends also on the neurons number *N*. In section 9.1 we have shown that the best value of $c_\beta$ is insensitive to other parameters, we can usually search the space $d-N$ for the best control parameter set by fixing $c_\beta$ at $c_\beta = 0.005$.

## 9.3 The role of neuron transfer functions and cost functions

For sake of the simplicity, we study here cost functions of $F_1$ and $F_2$, and transfer functions of the Gaussian and sigmoid types. Figure 14 shows the results with the combinations $F_1$-Gaussian (up-triangles), $F_1$-sigmoid (up-triangles), $F_2$-sigmoid (down-triangles), and $F_2$-Gaussian (circles), correspondingly. They are obtained at *N=3000*. The training stop conditions are $F_1 < 10^{-4}$ or $F_2 < 1$ correspondingly. Obviously, the cost function $F_2$ with the Gaussian transfer function gives the best result, indicating that Gaussian type functions march better the nature of the spatial pattern.

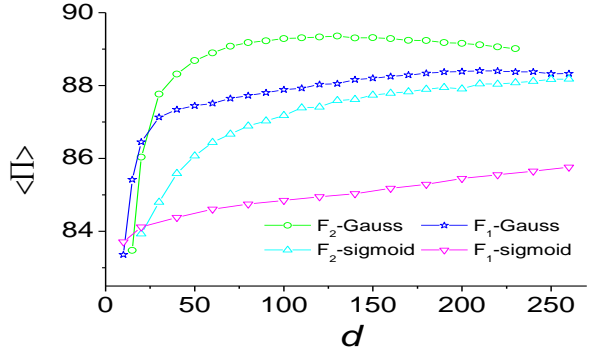

Figure 14. The dependence of the average recognition rate on transfer functions and cost functions.

## 9.4 Using a J-GVM

Figure 15 shows $\Pi$, $<\Pi>$ and $<\Pi^J>$ as functions of $d$. At each point of $d$, 500 $28\times28-3000-10$ GVMs are designed using the first 1% MNIST samples. Fig. 15(a) shows the results applying the Gaussian transfer function and Fig. 15(b) applying the polynomial transfer function with n=7. The cost functions are both Gaussian type.

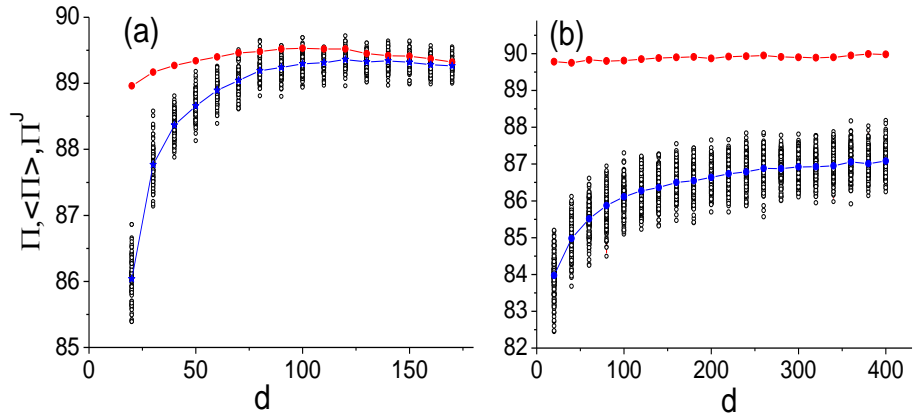

Figure 15. $\Pi$ ( circles), $<\Pi>$ ( solid stars) and $\Pi^J$ (solid circles) as functions of $d$, (a) for the Gaussian transfer function and (b) for the polynomial transfer function respectively.

One can see that, besides having a high value, the recognition rate is relatively insensitive to the control parameter. In Fig. 15(a) with the

Gaussian transfer function, J-GVMs designed in $d \in (50,150)$ all have approached recognition rate, and in Fig. 15(b) with the polynomial transfer function, the recognition rate of the J-GVM seems just slightly dependent on $d$. This property is also an essential advantage of a J-GVM since the carful-searching in the parameter space for the best control parameter set is avoided.

By using sufficient GVMs to construct the J-GVM, the risk can be suppressed by the ensemble average even individual GVMs having relatively big risk. This is why the J-GVM has high recognition rate and is insensitive to the control parameter. Fig. 15(a) also shows another effect, i.e., the best control parameter $d$ for the optimal J-GVM is smaller than that for the optimal GVMs. As will also seeing in next section, this is a common property for a-GVM. The reason may be as follows. The recognition rate is determined by features being extracted from the samples. Different GVM extracts information from different angle. GVMs at their best control parameter set have relatively small design risk, and thus have relatively small dispersion in 'angles'. On the contrary, GVMs with relatively big input-output sensitivity extract feature information from more wide 'angles'.

It is interesting in applying the polynomial transfer function. Though the recognition rates of GVMs are quite low, the rate of the J-GVM is even higher than that of using the Gaussian transfer function. The reason may be that this transfer function marches the nature of digit patterns better. Because the change of the gray degree of digit patterns is steep, higher-order polynomial transfer functions fit this feature well. Nevertheless, because $f_i^{"}(z) = n(n-1)z^{n-2}$ the structural risk of neurons is remarkably big, individual GVMs with the polynomial transfer function have bad performance. By applying a J-GVM, the risk is suppressed by the ensemble average, and the advantage that the high-order polynomial transfer functions emerges.

## 9.5 Improving the performance further by proper pretreatment of samples

As explained in section 9.1, the maximum-margin strategy is generally applicable when the test samples can be considered as random variations of training samples. For handwritten digits as well as usual spatial patterns, the variations could not be considered as random. The particular geometry of spatial patterns excludes most of the random variations. To create a virtual sample set following the geometric nature of patterns is a way to avoid the excessive generalization, and the tangent distance technique [18], involving shift, distoration, rotation, etc. can be used for this purpose. We have constructed a spurious sample set in section 9.1 by shifting each of the first 1% MNIST sample patterns with 2 pixel units to adjacent positions. It is applied as a test set there. Here we apply it to be a training set. Figure 16 shows that the recognition rate on the test set is dramatically improved comparing to that using the original 1% samples. The main drawback is that the training will be time-consuming when the amount of samples is too large.

Certain simple retreatments of input vectors may also effective. For example, we smooth the 1% samples by using a Gaussian convolution with unit standard deviation and applied them to train the learning machine, the recognition rate is also improved, see Fig. 16. The more effective way of encoding the spatial information is the Gradient-based feature extracting technique developed in recent years [19]. A 200-dimensional numeric feature vector encoding eight direction-specific $5 \times 5$ gradient images is calculated for each sample using this technique. This is one of three top-performing representations in [19] and is called

*e-grg* in their paper. Applying the first 1% samples pretreated by this technique we obtain a much high recognition rate, as shown in Fig. 16. Note that we do not apply the 1% samples randomly choosing from the whole training set to be the training samples, as done in ref. [19]. In that way one does not know which samples actually been chosen and thus may hinder a fair comparison among different researchers.

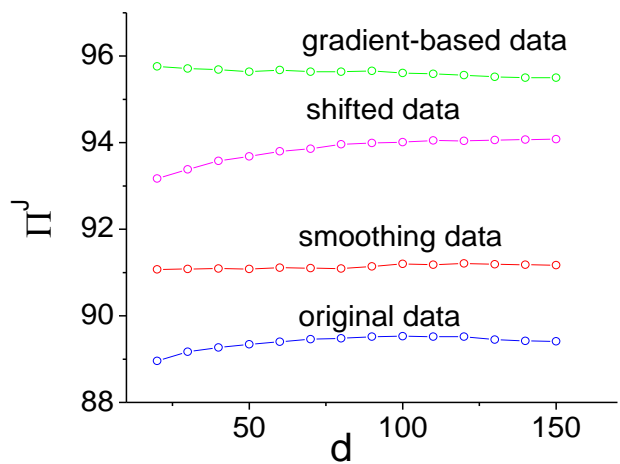

Figure 16. The recognition rate as a function $d$ at $c_\beta = 0.005$ and $N = 3000$ for the J-GVM designed by different training set constructed by the original 1% MNIST data set.

## 9.6 The highest record on the data set

Similar to other training methods of learning machine, increasing training samples can increase the recognition rate. The superiority of the GVM may become weaken when the amount of samples is sufficiently large, like other methods. Fig. 17 (a) and Fig. 17(b) show the results of using the first 10% and all of the MNIST samples respectively. In the first case 50 GVMs, and in the last case 10 ones, are trained respectively at each parameter point, and the J-GVM is constructed with these GVMs. In both cases, the Gaussian transfer function is applied and the GVM size is fixed at N = 6000. The cost function $F_2$ with training termination condition $F_2 < 1$ is applied. In the training, the normalized gray-scale images are directly used so as to purely compare the algorithms themselves, with getting rid of the improvement resulted by pretreatment techniques. It can be seen that using all of the training samples the record is beyond those using the BP method with error rate 1.5% [20], the SVM with error rate 1.4% (By combining 10 one-vs-rest binary SVMs and building a ten-class digit classifier) [21] and the recently improved deep-learning method with error rate 1.25%[22]. The last record is obtained with a complex five-layer hierarchical model. Therefore, in the case of applying the original training set, our record is competitive. Inspire this is the case, we still emphasize that the priority of our method is particularly for small sets of samples.

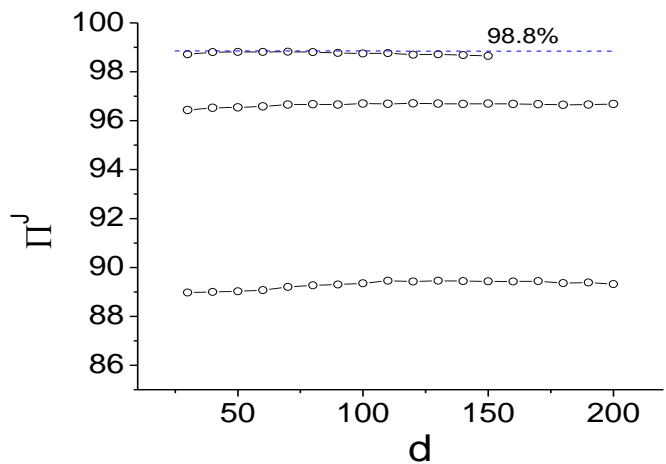

Figure 17. The recognition rate as a function $d$ at $c_\beta = 0.005$ and $N = 6000$ for the J-GVM designed by the original first 1%, 10%, and the complete set of MNIST data set, correspondingly.

## 10 Classification

Classification is a special case of pattern recognition. The Wisconsin breast cancer database was established [14] at 1992 with 699 samples. As usual, the first 2/3 samples are applied as the training set, and the remains as the test set. The inputs are 9 dimensional vectors, with components represent features from microscopic examination results, and are normalized to take value from [0, 10]. The output indicates the benign and malignant patients.

This task can be achieved by a GVM with two neurons in the output layer. A patient is classified into benign if the output of the first neuron is bigger than that of the second one, otherwise malignant. We first study $<\Pi>$ and $E[\Pi]$ as a function of the control parameter $d$ with other parameters keeping fixed at $\beta_i \in [-1,1]$, $\overline{w}_{jk} \in [-1,1]$, $b_i \in [-10,10]$, $\alpha_i = 1$, $w_{il} = \pm 1$ and $N=200$. At each set of control parameters, 500 GVMs are used to calculate $<\Pi>$ and $E[\Pi]$.

Figure 18 shows the results. The representations of the symbols are: up-triangle for $F_1$- sigmoid combination; down-triangle for $F_1$-Gaussian combination; star for $F_2$-sigmoid combination; circles for $F_2$- Gaussian combination. The stop conditions are $F_1 < 10^{-3}$ or $F_2 < 1$ correspondingly. When the stop conditions cannot be fulfilled within a preset maximum training time, the search alone $d$ is ceased.

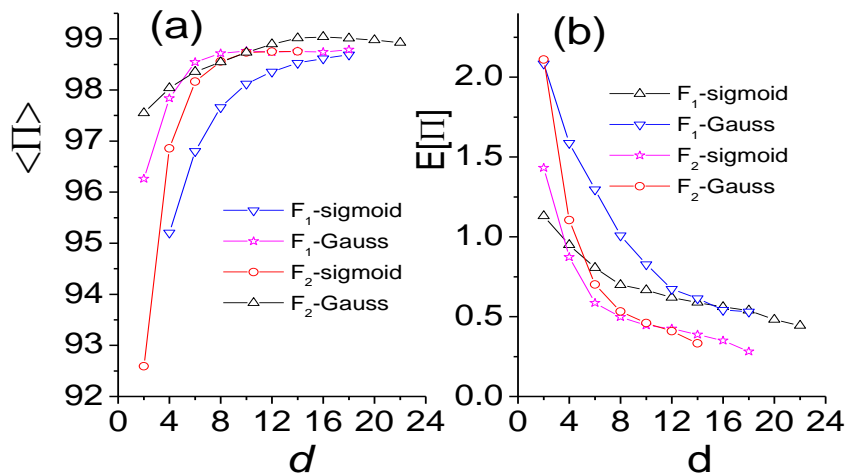

Figure 18 $<\Pi>$ and $E[\Pi]$ as a function of $d$.

It can be seen that the best result is given by the steep cost function $F_1$ with the sigmoid transfer function. This fact indicates that steep functions march the nature of this sample set well. The reason is that for a component of such a sample vector, small value means the normal, while large one represents the abnormal.

With the $F_1$ - sigmoid combination, the maximum of $<\Pi>$ is approached around $d \sim 16$, after which it become decrease slightly. This phenomenon might be explained as that the data from the medical examination could be regarded approximately as random variations of 'standard samples'. The microscopic examination may induce random errors of measurement, and meanwhile biochemical indexes themselves may be influenced in a complex way by prompt accidental events of a patient.

Fig. 18(b) shows that $E[\Pi]$ decreases monotonously with the increase of $d$. The essential feature explored here is the big uncertainty. For example, in using the $F_1$-sigmoid combination, $<\Pi^{GVM}>$ is about $\pm 0.5\%$ even at $d=16$ where the average correct rate takes the maximal value of $99.0\%$. That is to say, different GVM obtained by different user following the same training program may show correct rate region from $98.5\% - 99.5\%$. The uncertainty is thus a serious problem. This drawback is a consequence of small sample set.

Applying a J-GVM is the effective way to overcome this drawback. Fig. 19 shows the distribution of the recognition rate of GVMs as a function of the control parameter $d$. At each point of $d$, the correct rates of 500 GVMs designed with the $F_1$-sigmoid combination are shown as stars. The average correct rate of GVMs and the correct rate of the J-GVM constructed using these GVMs are also shown in the figure as triangles and circles respectively. Fig. 19(a) is for $N=200$ and Fig. 19(b) for

*N=500*. In the case of N=200, the maximal average correct rate is 99.01%. In an interval of $d \in [4,8]$, the rate of the J-GVM keeps at 100%. With more big GVMs, N=500, the maximum average correct rate approaches 99.30% and in a wide interval of $d \in [4,22]$ the correct rate of the J-GVM keeps at 100%. This fact again indicates that big machines are more favorable. As to the records of the correct rate, our results are superior to previous studies [14-15], where a record of 97.5% is approached by a SVM-method based learning machine.

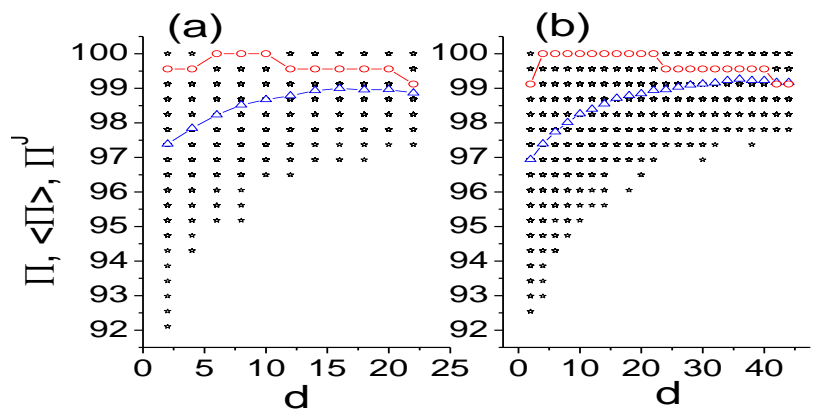

Figure 19. $\Pi$ ( circles), $<\Pi>$ ( solid stars) and $\Pi^J$ (solid circles) as functions of *d*, (a) for N=200 and (b) for N=500.

Figure 19 also explores the drawback of applying the individual learning machine to be the performing machine. Even in the parameter region with low average correct rate, as at *d*= 2 in Fig. 19, certain GVMs may give the correct rate of 100%. However, at the same parameter, another one may approach only about 92%. Therefore, the correct rate on the test set is not a good indicator variable of the learning machine performance. The higher record may be just a fluctuation. When applying it to real patients, one cannot expect it still remain the correct rate. Indeed, the learning machine with 92% record may not be necessarily worse than that with 100% record for real application. The J-GVM can avoid such kind of uncertainty.

# 11 Summary and Discussion

(1) We develop a MC algorithm to gain the correct response to the training set. The basic idea is to drive the local fields of neurons move continuously towards the target distribution defined by the cost function. Applying this algorithm, one can obtain three-layer neural networks with either continuous or discrete parameters. The MC algorithm works well mainly because each random adaptation is performed only for one parameter in the hidden layer, and thereby it results only $O(P + LP)$ multiply operations for making the decision, other than $O(NMP + NLP)$ operations required for evolving the whole network.

Comparing to the SVM method, we give up support vectors, and replace them by general weight vectors. Support vectors are special weight vectors. For small training-set problems, support vectors are limited by the number of samples. The weight vectors have no such a limitation. Using enough weight vectors, the features of input vectors can be maximally extracted

(2) We classify the prior knowledge into common and problem-dependent parts, and suggest corresponding strategies to incorporate them into the learning machine to gain maximum generalization ability. There are two classes of common prior knowledge. The first part is that the learning machine should not be too sensitive to the small changes of inputs. This is resulted by the prior knowledge that normal functions usually have sufficient smoothness, and variations of a knowing pattern may belong to the same class. The second part is a basic requirement for a design method itself. Following the same rule, the same specified initial condition, and the same training set, learning machines designed by different users should have sufficient small dispersion on the same test set. This requirement usually does not emphasized in previous training methods. Here we apply it as a basic principle to supervise the design of learning machines. This is the DRM strategy. We have argued that the design risk can be saved as the unique quantitative indicator variable of the best control parameter set for function approach and smoothing, which has been confirmed by examples. On the contrary, we illustrated that the SRM principle may induce over-minimization of the risk and induce the deviation from the best fitting. We have also demonstrated that minimizing the design risk can lead to the maximization of the separation margin for the classification, and thus can gain the better generalization ability. The maximum margin strategy, however, is exactly available when the real patterns can be considered as random variations of the sample patterns. In other word, it may induce over training for certain pattern samples, such as the handwriting digits. Real patterns usually have particular geometric symmetry and extremely maximizing the separating margin may result the mismatch to the nature symmetry.

Therefore, it is essential for further improving the generalization ability by incorporating the problem-dependent prior knowledge. For function approach and smoothing, choosing a more proper neuron transfer function is such a way. The more proper function can be also chosen using the design-risk criterion. For pattern recognition and classification, there are many manners to maximize the problem-dependent prior knowledge, such as using a more proper neuron transfer function or cost function. One can also construct spurious samples following the particular geometric symmetry of samples to extend the training set, or incorporate the particular geometric information into input vectors. In this case, we need to combine the average recognition rate with the design risk to find the best control parameter set. To calculate the average correct rate, the real test set is not must necessary. One may construct a spurious sample set having the same geometric symmetry to be the test set.

As a result of the DRM strategy, instead of finding the best machine according to the test result on the test set, we search for the best control parameter set. At the best control parameter set, each GVM is equivalent to applications and can be applied as the performing leaning machine.

(3) The structural risk is still a key parameter in our method. We usually search for the best control parameter set alone the direction with decreasing input-output sensitivity. However, we control the structural risk of a GVM by control that of individual neurons. It is show that the structural risk of a neuron is determined by the multiple of several classes of parameters as well as the second derivative of the neuron transfer function. By limiting the intervals of these parameters and applying proper neural transfer functions which second derivatives have fixed upper bound, the risk of individual neurons is controllable. As a linear combination of the risks of individual neurons, the structural risk of the machine thus can be controlled.

(4) We can apply the J-GVM constructed by a sufficient amount of GVMs to be the performing machine. The output of a J-GVM is the ensemble average of these GVMs. It is an application of the ensemble strategy [11]. The J-GVM usually has better performance since it has more small empirical risk, structural risk and design risk. The flexibility of the Monte Carlo algorithm enables us to obtain a sufficient amount of statistically identical GVMs using the same training set so as to effectively diminish the noise part.

(5) We emphasize that the superiority of our method is for small sample-set problems. For function approach and smoothing, examples show that the fitting precision using small training set is obviously higher than those using the SVM method and the conventional spline algorithm.

In the example of classification of breast cancer patients using totally 699 samples, the result is encouraging. Particularly with a J-GVM one can always get the 100% correct rate in a wide region of control parameters, indicating the learning machine has a high confidence level.

For handwritten digit recognition, historical records of the highest recognition rate may result confusion since they have complex background. The training samples may be pretreated using various techniques and the learning machine may have multilayer structures more than the three. In our paper we apply only the original gray-scale images without any pretreatment so as to fairly compare the method itself. We approach a recognition rate of 90% by use of only the first 1% samples, and of 97% by use of the first 10% samples. Using all 60000 samples, we obtained the recognition rate of 98.8%. This rate is beyond those using the BP method with error rate 1.5% [20], the SVM with error rate 1.4% [21] and the recently improved deep-learning method with error rate 1.25%[22]. For these record, only the BP method is obtained with a three-layer neural network. The record of the SVM is for using 10 bi-classifiers. If applying the multi-classifier SVM, only a correct rate of 96% is approached. The last record is obtained with a complex five-layer hierarchical model. Inspire this is the case, we still emphasize that the priority of our method is particularly for small sets of samples. In fact, when the training samples are sufficient, the SVM method also has great freedom to select support vectors, which reduces the superiority of our strategy.

(6) Our method can applied to other problems. The method can be applied directly on many other traditional tasks of leaning problem, such as the time-series prediction. What is more, the algorithm may have induce some particular applications. For example, after proceeding the Monte Carlo adaptation for a proper period, the local fields $h_i^\mu s_i^\mu$ will distribute around $s_i^{\mu} h_i^{\mu} = d$. Then those examples with much small $h_i^{\mu} s_i^{\mu}$ may represent 'bad examples', such as two identical objects with opposite labels. The figure below shows 20 such examples. One can see that, the third sample and the last one for example, no one could be recognize them as '3' and '4' respectively. To pick out these bad examples, the test set is not used. It may be an instance of the so-called transductive inference [4]. Picking up bad examples might have practical importance for certain problem, such as finding those misdiagnosed patients from the training set.

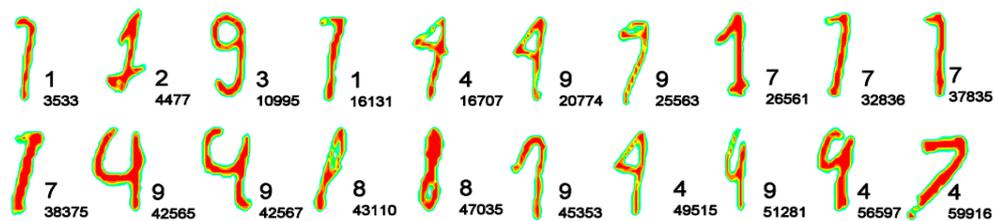

Figure 20, 'Washing out' the bed samples. The digit patterns are from the MNIST data set. The above numbers of the subscript are target digits, below ones are the sequence number of patterns in the data set, correspondingly

As a new attempt of developing the design theory of learning machines, many viewpoints are presented without strict proof. Nevertheless, as a research field to solve practical problems, suggesting new algorithms based on empirical study and then investigating their theoretical basis, is consistent with common practices of this research field.